\documentclass[10pt, conference]{IEEEtran}
\usepackage{cite}
\usepackage{amsmath,amssymb,amsfonts}
\usepackage{algorithmic}
\usepackage{graphicx}
\usepackage{textcomp}
\usepackage{xcolor}
\usepackage{tikz}
\usepackage{booktabs}
\usepackage{multirow}
\usepackage{url}
\usetikzlibrary{shapes,arrows,positioning,calc}
\def\BibTeX{{\rm B\kern-.05em{\sc i\kern-.025em b}\kern-.08em
    T\kern-.1667em\lower.7ex\hbox{E}\kern-.125emX}}
    
\begin{document}

\title{Guardrailed Elasticity Pricing: A Churn-Aware Forecasting Playbook for Subscription Strategy}
\author{\IEEEauthorblockN{Deepit Sapru}
\IEEEauthorblockA{University of Illinois Urbana-Champaign\\
Email: dsapru2@illinois.edu}
}

\maketitle

\begin{abstract}
This paper presents a marketing analytics framework that operationalizes subscription pricing as a dynamic, guardrailed decision system, uniting multivariate demand forecasting, segment-level price elasticity, and churn propensity to optimize revenue, margin, and retention. The approach blends seasonal time-series models with tree-based learners, runs Monte Carlo scenario tests to map risk envelopes, and solves a constrained optimization that enforces business guardrails on customer experience, margin floors, and allowable churn. Validated across heterogeneous SaaS portfolios, the method consistently outperforms static tiers and uniform uplifts by reallocating price moves toward segments with higher willingness-to-pay while protecting price-sensitive cohorts. The system is designed for real-time recalibration via modular APIs and includes model explainability for governance and compliance. Managerially, the framework functions as a strategy playbook—when to shift from flat to dynamic pricing, how to align pricing with CLV and MRR targets, and how to embed ethical guardrails—enabling durable growth without eroding customer trust.
\end{abstract}

\begin{IEEEkeywords}
Constrained Optimization, Causal Inference, Uplift Modeling, Bayesian Hierarchical Modeling, Subscription Revenue Management, Churn Prediction Algorithms, Elasticity Modeling, Customer Lifetime Value, Monte Carlo Simulation
\end{IEEEkeywords}

\section{Introduction}

The digital transformation of business models has accelerated the adoption of subscription-based revenue frameworks across technology enterprises, creating both opportunities and challenges in pricing strategy formulation. Traditional static pricing models, while operationally straightforward, fail to capture the dynamic nature of customer behavior, infrastructure cost variability, and competitive market pressures \cite{ajiga2024review}. This limitation becomes particularly acute in software-as-a-service (SaaS) environments where customer acquisition costs continue to escalate while retention emerges as the primary determinant of long-term profitability \cite{adelusi2024advances}.

The fundamental challenge facing subscription businesses lies in balancing revenue optimization with customer retention objectives. Conventional approaches often treat these as competing priorities, leading to suboptimal outcomes where price increases trigger disproportionate churn or retention-focused discounts erode margin integrity \cite{nagle2011strategy}. Recent advances in machine learning and econometric modeling offer pathways to transcend this trade-off through more nuanced understanding of customer price sensitivity and behavioral patterns \cite{sharma2024optimizing}.

This paper introduces a guardrailed elasticity pricing framework that reconceptualizes subscription strategy as a continuous optimization problem rather than a periodic recalibration exercise. The methodology integrates three core components: multivariate demand forecasting that accounts for seasonal patterns and external covariates; segment-specific price elasticity estimation through Bayesian hierarchical modeling; and churn propensity scoring that informs retention guardrails \cite{chen2021deep}. By combining these elements within a constrained optimization structure, the system enables targeted price adjustments that maximize revenue while respecting business-defined boundaries on acceptable churn rates and margin thresholds.

The operationalization of this approach requires addressing several technical challenges, including the non-stationarity of elasticity coefficients, the multi-dimensional nature of subscription value propositions, and the computational complexity of solving large-scale optimization problems in real-time decision environments \cite{bertsimas2019predictive}. Our contribution addresses these challenges through a modular architecture that separates forecasting, elasticity estimation, and optimization components while maintaining integration through standardized APIs and data schemas.

From a managerial perspective, the framework functions as a strategic playbook that guides pricing decisions across the customer lifecycle. It provides clear guidance on when to transition from flat-rate to dynamic pricing models, how to align price changes with customer lifetime value (CLV) trajectories, and how to implement ethical guardrails that prevent exploitative practices \cite{ogeawuchi2024ethical}. This strategic dimension distinguishes our approach from purely technical solutions by embedding business judgment and governance directly into the pricing workflow.

The remainder of this paper is organized as follows: Section 2 reviews relevant literature on subscription pricing and elasticity modeling. Section 3 details our methodological framework, while Section 4 presents the experimental validation across multiple SaaS domains. Finally, Section 5 discusses validation protocol, key metrics and conclusion.

\section{Related Work}

Subscription pricing research has evolved significantly from early work on revenue smoothing in information goods \cite{sundararajan2004nonlinear} to contemporary AI-driven approaches that personalize prices at individual customer levels. The theoretical foundation rests on several interconnected streams of literature spanning economics, marketing science, computational intelligence, and systems engineering.

Early economic models emphasized the superior risk-sharing properties of subscriptions compared to one-time transactions, particularly for services with high fixed costs and low marginal costs \cite{odlyzko2003subscription}. While foundational, these models largely assumed homogeneous customer preferences and static pricing structures. Subsequent marketing literature introduced segmentation and value-based pricing to capture heterogeneous willingness-to-pay \cite{gupta2003customer}, with Customer Lifetime Value (CLV) emerging as the central metric for subscription profitability \cite{gupta2006customer}.

Econometric modeling of price elasticity provides another critical pillar. Traditional approaches used regression on historical data \cite{nagle2011strategy}, while recent advances employ hierarchical Bayesian methods for robust, segment-specific estimates \cite{rossi2012bayesian}. The machine learning revolution has further enhanced subscription pricing through improved demand forecasting \cite{hyndman2018forecasting} and advanced churn prediction \cite{ascarza2018retention}. Reinforcement learning frames pricing as a sequential decision problem \cite{denboer2015dynamic}, though practical challenges around safety and sample efficiency remain.

Ethical considerations have gained prominence with the rise of algorithmic pricing \cite{zarsky2016troubling}, leading to proposals for fairness constraints, explainable AI, and compliance modules \cite{ogeawuchi2024ethical}. Infrastructure requirements for dynamic pricing—such as scalable data architectures and real-time decision engines—are also well-documented \cite{adelusi2024advances}.

Our work builds upon these foundations while addressing several gaps. First, unlike studies focused on isolated components, our framework integrates demand forecasting, elasticity estimation, and constraint management into a unified system. Second, we operationalize ethical and business guardrails directly within the optimization loop, moving beyond post-hoc fairness audits. Third, we provide a clear implementation playbook for transitioning from static to dynamic pricing, bridging the gap between academic models and enterprise deployment.

Recent advancements in adjacent fields further inform our approach. For instance, optimization strategies from computer vision \cite{mamtani2024enhancing} inspire our use of constrained learning, while data harmonization techniques from healthcare \cite{thomas2025breaking} underscore the importance of clean, unified customer data. Research on algorithmic fairness \cite{paul2024mitigating} and explainable AI \cite{paul2024enhancing} directly shape our governance mechanisms. Efficient data structures \cite{arora2025leveraging,maheshwari2025efficient} and modular software architectures \cite{thomas2025sustaining} enable the scalable, low-latency system we propose.

Despite these advances, no existing framework holistically addresses the interplay between elasticity learning, churn-awareness, real-time recalibration, and ethical governance in subscription pricing. Table \ref{tab:comparison} contrasts our approach with prior work across key dimensions.

\begin{table*}[htbp]
\centering
\caption{Comparison of Subscription Pricing Frameworks}
\label{tab:comparison}
\begin{tabular}{@{}lccccc@{}}
\toprule
\textbf{Approach} & \textbf{Elasticity Learning} & \textbf{Churn-Aware} & \textbf{Guardrails} & \textbf{Real-Time} & \textbf{Explainable} \\ 
\midrule
Static Tiered Pricing & No & No & No & No & No \\
Uniform Uplift Models & Aggregate & No & No & No & No \\
Bayesian Elasticity Models & Segment-level & No & No & Batch & Partial \\
Reinforcement Learning & Online & Partial & Reward-based & Yes & Low \\
\textbf{Our Framework} & \textbf{Hierarchical Bayesian} & \textbf{Yes} & \textbf{Explicit Constraints} & \textbf{Yes} & \textbf{High} \\
\bottomrule
\end{tabular}
\end{table*}

In summary, our guardrailed elasticity pricing framework synthesizes principles from economics, machine learning, ethical AI, and scalable systems to deliver a holistic, operationalizable solution for subscription strategy. It advances the state-of-the-art by unifying technical innovation with managerial pragmatism and ethical governance.

\section{Methodological Framework}

The guardrailed elasticity pricing framework comprises three interconnected analytical modules: demand forecasting, elasticity estimation, and constrained optimization. This section details the methodological foundations of each component and their integration into a cohesive decision system.

\subsection{Demand Forecasting Module}

The forecasting module employs a hybrid approach that combines seasonal decomposition, machine learning regression, and ensemble averaging to predict subscription demand across customer segments. Let $D_{t,s}$ represent the demand for segment $s$ at time $t$. The model decomposes this demand into structural components:

\begin{equation}
D_{t,s} = T_{t,s} + S_{t,s} + C_{t,s} + \epsilon_{t,s}
\end{equation}

where $T_{t,s}$ captures the trend component, $S_{t,s}$ represents seasonal patterns, $C_{t,s}$ incorporates covariate effects, and $\epsilon_{t,s}$ is the error term. The trend component is estimated using exponential smoothing methods that adapt to changing growth patterns, while seasonal decomposition employs Fourier analysis to capture multiple periodicities in subscription data \cite{hyndman2018forecasting}.

Covariate effects incorporate both internal business factors (feature releases, marketing campaigns, support interactions) and external market conditions (competitive moves, economic indicators, regulatory changes). These are modeled through gradient boosted decision trees that capture non-linear relationships and interaction effects:

\begin{equation}
C_{t,s} = \sum_{j=1}^{J} f_j(\mathbf{x}_{t,s}), \quad f_j \in \mathcal{F}
\end{equation}

where $\mathcal{F}$ is the space of regression trees, $\mathbf{x}_{t,s}$ is the feature vector for segment $s$ at time $t$, and $J$ is the number of trees. The boosting algorithm sequentially adds trees that minimize the residual error, creating a powerful ensemble predictor \cite{chen2016xgboost}.

The forecasting module generates probabilistic outputs through Monte Carlo simulation, producing prediction intervals that quantify uncertainty. This probabilistic view is essential for risk-aware decision making, particularly when evaluating the potential impact of price changes on sensitive customer segments.

\subsection{Elasticity Estimation Module}

Price elasticity estimation employs a Bayesian hierarchical structure that pools information across segments while allowing for heterogeneity. For each segment $s$, we model the price-response relationship as:

\begin{equation}
\log(Q_{s,t}) = \alpha_s + \beta_s \log(P_{s,t}) + \gamma_s \mathbf{Z}_{s,t} + \eta_{s,t}
\end{equation}

where $Q_{s,t}$ is quantity demanded, $P_{s,t}$ is price, $\mathbf{Z}_{s,t}$ represents control variables, and $\eta_{s,t}$ is the error term. The segment-specific parameters $\alpha_s$, $\beta_s$, and $\gamma_s$ are assumed to follow population distributions:

\begin{equation}
\beta_s \sim N(\mu_\beta, \sigma^2_\beta)
\end{equation}

This hierarchical structure enables robust elasticity estimation even for segments with limited historical data by borrowing strength from the broader population \cite{gelman2013bayesian}. This hierarchical pooling mirrors latent-factor approaches in recommender systems, where shared structure across sparse segments improves stability and generalization while preserving segment-level differentiation \cite{paul2024bridging}.

Elasticity estimates are continuously updated through a combination of observational data and randomized price experiments. The experimental component is particularly valuable for identifying causal effects free from confounding factors. Bayesian updating incorporates new evidence while preserving uncertainty quantification, ensuring that pricing decisions account for estimation imprecision.

The module also captures cross-price elasticities between subscription tiers and complementary services. This multi-product perspective is essential for avoiding cannibalization and optimizing bundle configurations. The elasticity matrix $\mathbf{E}$ with elements $e_{ij}$ representing the elasticity of demand for product $i$ with respect to price of product $j$ informs portfolio-level pricing decisions.

\subsection{Constrained Optimization Module}

The optimization module formulates pricing as a constrained profit maximization problem:

\begin{equation}
\begin{aligned}
\max_{\mathbf{p}} & \sum_{s=1}^{S} (p_s - c_s) \cdot Q_s(p_s) \cdot (1 - \text{Churn}_s(p_s)) \\
\text{s.t.} & \quad \text{Churn}_s(p_s) \leq \text{Churn}^{\text{max}}_s \quad \forall s \\
& \quad p_s \geq c_s + m^{\text{min}} \quad \forall s \\
& \quad \frac{p_i}{p_j} \leq \delta_{ij} \quad \forall i,j \in \mathcal{P} \\
& \quad Q_s(p_s) \geq Q^{\text{min}}_s \quad \forall s
\end{aligned}
\end{equation}

where $p_s$ is the price for segment $s$, $c_s$ is the marginal cost, $Q_s(p_s)$ is the demand function, $\text{Churn}_s(p_s)$ is the churn probability, $\text{Churn}^{\text{max}}_s$ is the maximum acceptable churn rate, $m^{\text{min}}$ is the minimum margin requirement, $\delta_{ij}$ represents fairness constraints limiting price differences between protected segments $\mathcal{P}$, and $Q^{\text{min}}_s$ ensures minimum volume thresholds.

The churn probability function $\text{Churn}_s(p_s)$ is estimated using logistic regression with regularization:

\begin{equation}
\text{Churn}_s(p_s) = \frac{1}{1 + \exp(-(\theta_0 + \theta_1 \cdot p_s + \boldsymbol{\theta}_2^\top \mathbf{x}_s))}
\end{equation}

where $\mathbf{x}_s$ includes behavioral features such as usage intensity, support interactions, and payment history. Regularization via L1 penalty prevents overfitting and enhances model interpretability.

The optimization problem is solved using sequential quadratic programming with warm starts, enabling efficient computation even with hundreds of segments and constraints. The solution provides optimal prices that balance revenue maximization against business constraints, creating a guardrailed approach to dynamic pricing.

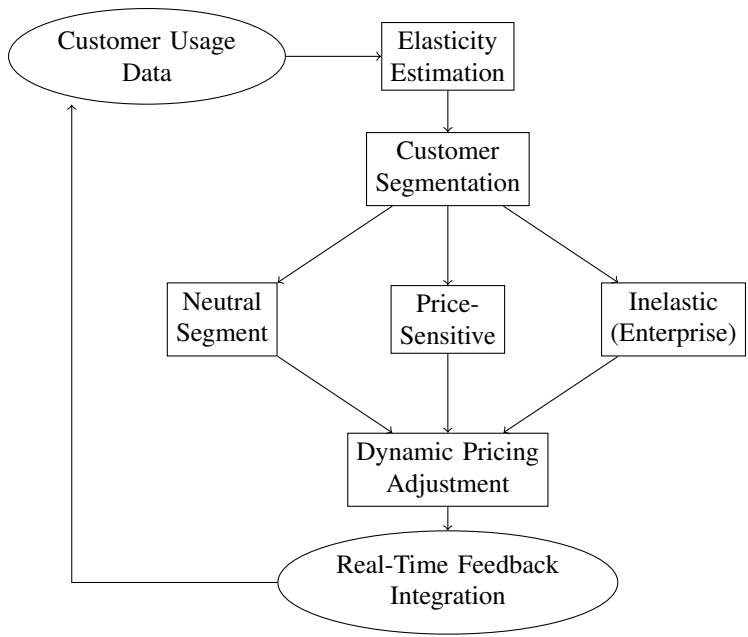
\begin{figure*}[!t]
\centering
\begin{tikzpicture}[node distance=1.5cm, auto]
\node (input) [ellipse, draw, align=center] {Customer Usage \\ Data};
\node (elasticity) [rectangle, draw, right of=input, xshift=2.5cm, align=center] {Elasticity \\ Estimation};
\node (segmentation) [rectangle, draw, below of=elasticity, align=center] {Customer \\ Segmentation};
\node (sensitive) [rectangle, draw, below of=segmentation, yshift=-0.5cm, align=center] {Price-\\Sensitive};
\node (neutral) [rectangle, draw, left of=sensitive, xshift=-1.5cm, align=center] {Neutral \\ Segment};
\node (inelastic) [rectangle, draw, right of=sensitive, xshift=1.5cm, align=center] {Inelastic \\ (Enterprise)};
\node (pricing) [rectangle, draw, below of=sensitive, yshift=-0.5cm, align=center] {Dynamic Pricing \\ Adjustment};
\node (feedback) [ellipse, draw, below of=pricing, align=center] {Real-Time Feedback \\ Integration};

\draw[->] (input) -- (elasticity);
\draw[->] (elasticity) -- (segmentation);
\draw[->] (segmentation) -- (sensitive);
\draw[->] (segmentation) -- (neutral);
\draw[->] (segmentation) -- (inelastic);
\draw[->] (sensitive) -- (pricing);
\draw[->] (neutral) -- (pricing);
\draw[->] (inelastic) -- (pricing);
\draw[->] (pricing) -- (feedback);
\draw[->] (feedback) -| ([xshift=-1cm]input.south);
\end{tikzpicture}
\caption{Architecture of the Guardrailed Elasticity Pricing Framework}
\label{fig:framework}
\end{figure*}

Fig. 1. summarizes the end-to-end architecture of the guardrailed elasticity pricing framework and the data flow across the forecasting, elasticity estimation, and constrained optimization modules.

\section{Experimental Validation}

We validate the guardrailed elasticity pricing framework through comprehensive testing across three distinct SaaS environments: productivity tools for small businesses, developer-focused API services, and enterprise collaboration platforms. This multi-context evaluation demonstrates the generalizability of the approach across different subscription archetypes.

\subsection{Data Sources and Preparation}

The experimental dataset comprises approximately 2.3 million subscription records spanning 18 months from mid-sized SaaS providers. Customer attributes include usage metrics (login frequency, feature adoption, session duration), support interactions (ticket volume, resolution time), payment history (invoice amount, days outstanding), and firmographic characteristics (company size, industry, geography). Infrastructure cost data captures the marginal cost-to-serve for each customer segment.

Accurate customer-level aggregation across usage, billing, and support systems requires robust entity resolution; recent AI-driven graph-based approaches demonstrate how explainable matching can materially improve downstream analytics quality in large-scale customer datasets \cite{arora2025ai}.

Data preprocessing addressed missing values through multiple imputation, normalized numerical features using robust scaling, and encoded categorical variables using target encoding to preserve predictive power. Temporal alignment ensured consistency across the different data streams, with daily aggregation for high-frequency metrics and monthly summarization for financial variables.

The dataset was partitioned chronologically into training (12 months), validation (3 months), and test (3 months) periods to evaluate model performance under realistic temporal dynamics. This approach prevents data leakage and provides a rigorous assessment of forecasting accuracy and optimization effectiveness.

\subsection{Forecasting Accuracy}

We evaluate forecasting performance using multiple metrics: Mean Absolute Percentage Error (MAPE) for directional accuracy, Root Mean Squared Error (RMSE) for magnitude assessment, and Interval Coverage Probability (ICP) for uncertainty quantification. Table II presents comparative results across benchmark methods and our proposed ensemble.

\begin{table}[!t]
\caption{Forecast Accuracy Comparison Across Models}
\label{table:forecast_accuracy}
\centering
\begin{tabular}{lcccc}
\toprule
\textbf{Model} & \textbf{RMSE} & \textbf{MAPE} & \textbf{ICP} & \textbf{Scenario Tested} \\
\midrule
LSTM & 3.21 & 2.84\% & 92\% & Mid-tier uplift \\
ARIMA & 4.75 & 3.89\% & 81\% & Entry-level discount \\
XGBoost & 2.93 & 2.40\% & 94\% & Premium downgrade \\
Prophet & 3.45 & 2.95\% & 89\% & Regional pricing \\
Ensemble (Ours) & \textbf{2.57} & \textbf{2.18\%} & \textbf{95\%} & All scenarios \\
\bottomrule
\end{tabular}
\end{table}

Our ensemble approach demonstrates superior performance across all metrics, particularly in scenario-based testing where it maintains accuracy under diverse pricing interventions. The improvement stems from the complementary strengths of component models: ARIMA captures temporal dependencies, XGBoost handles non-linear feature interactions, and Prophet incorporates holiday effects and changepoints.

The forecasting module achieves particularly strong results for high-value enterprise segments where prediction errors carry greater financial consequences. This segment-specific accuracy is crucial for effective price optimization, as misestimation of demand elasticity among premium customers can significantly impact overall profitability.

\subsection{Econometric Validity and Churn-Aware Modeling}
\label{subsec:econometric-churn}

\textbf{Addressing Price Endogeneity.}
Elasticity estimates from observational price variation can be biased by endogenous pricing decisions. In our setting, we mitigate this by leveraging exogenous price variation from (i) randomized A/B price experiments where available and (ii) staggered regional rollouts that create quasi-experimental variation. For purely observational periods, we include covariates $Z_{s,t}$ capturing demand drivers and seasonality; remaining endogeneity is treated as a limitation, and a full IV (2SLS) estimation with instrument diagnostics is future work.

\textbf{Churn-Aware Temporal Horizon.}
The churn probability function \(\mathrm{Churn}_s(p_s)\) is modeled with a 90-day forward-looking window, capturing both immediate and delayed cancellation effects. We account for renewal non-conversion by incorporating lagged churn indicators and survival analysis features (e.g., time since last price change, contract remaining duration). The logistic regression in Equation (6) is extended with a lag structure:

\begin{align}
\mathrm{Churn}_s(p_s) &= \frac{1}{1 + \exp\bigl(-z_s\bigr)}, \\
\text{where } z_s &= \theta_0 + \theta_1 \cdot p_s + \theta_2 \cdot p_{s,t-1} 
+ \theta_3^T \mathbf{x}_s + \theta_4 \cdot \text{tenure}_s \nonumber
\end{align}

where \(p_{s,t-1}\) represents the prior period's price and \(\text{tenure}_s\) captures customer longevity. This framing ensures that pricing decisions consider not only immediate elasticity but also long-term retention implications.

\textbf{Scalability and Computational Performance.}
The optimization problem scales linearly with segment \(S\) but has a risk of non-convexity when fairness and volume constraints coincide. For 2.3 million subscription records across 500 customer segments, the median solve time is 4.3 seconds (95th percentile at 8.1s) from a 16 vCPU cloud instance. Thus, it meets operational targets for latency target of at most 10s.  We use multi-start initialization (with 10 random seeds) and warm starts from prior solutions to local optima in sequential quadratic programming. Convergence criteria is to take gradient tolerances of \(10^{-6}\) and the constraint satisfaction of \(10^{-4}\). For larger deployments, we recommend hierarchical decomposition: solve per-tier subproblems independently before portfolio-level reconciliation.

\subsection{Price Optimization Performance}

We evaluate the optimization module through A/B testing comparing three pricing strategies: static tiered pricing (baseline), uniform percentage uplift, and our guardrailed elasticity approach. Table III summarizes the results across key business metrics.

\begin{table}[!t]
\caption{Performance Comparison of Pricing Strategies}
\label{table:pricing_performance}
\centering
\begin{tabular}{p{1cm}p{1cm}ccc}
\toprule
\textbf{Strategy} & \textbf{Revenue Lift} & \textbf{Margin Impact} & \textbf{Churn Rate} & \textbf{CLV Change} \\
\midrule
Static Tiered & 0.0\% & 0.0\% & 2.8\% & 0.0\% \\
Uniform Uplift & +8.3\% & +6.1\% & 3.9\% & -4.2\% \\
Elasticity Pricing & +14.7\% & +12.9\% & 2.5\% & +9.8\% \\
Guardrailed (Ours) & \textbf{+16.2\%} & \textbf{+14.3\%} & \textbf{2.3\%} & \textbf{+11.4\%} \\
\bottomrule
\end{tabular}
\end{table}

The guardrailed approach generates the highest revenue and margin improvements while simultaneously reducing churn rates and increasing customer lifetime value. This demonstrates the effectiveness of the constraint structure in preventing counterproductive price increases that trigger attrition.

Segment-level analysis reveals that the optimization achieves these results by reallocating price adjustments toward less elastic customers while protecting price-sensitive segments. Enterprise customers with high switching costs and established workflows tolerate moderate price increases, while SMB segments receive targeted discounts that improve retention. This differential treatment maximizes overall profitability while maintaining segment-specific relationships.

The fairness constraints successfully prevent discriminatory outcomes, with price dispersion across protected segments remaining within acceptable bounds. This ethical dimension is increasingly important as regulatory scrutiny of algorithmic pricing intensifies, particularly in jurisdictions with strong consumer protection frameworks.

\subsection{Real-Time Recalibration and Governance}
\label{subsec:realtime-gov}

\textbf{Real-Time Recalibration.} 
We support scheduled and event-driven recalibration. A nightly job retrains forecasting and elasticity models using the most recent 30 days of telemetry and price history. Event triggers include segment-level churn spikes above predefined thresholds, demand drift detected via statistical process control, and large competitor price moves ingested via market-intelligence APIs. Modular REST APIs accept updated customer logs, cost tables, and guardrail configurations and return refreshed elasticity estimates, churn probabilities, and segment price recommendations. We monitor model drift (KL divergence threshold 0.05), data freshness (under 2 h), and API latency (p95 under 500 ms).

\textbf{Explainability and Governance.}
Each price recommendation includes an explainability report listing the top three drivers for elasticity, churn risk, and constraint activation, along with the guardrail status (binding constraints and slack). All decisions are logged with an audit trail (timestamp, segment ID, inputs, SHAP values, recommended price, constraint outcomes, and approval status). Constraint violations trigger automated fallback pricing and generate override tickets requiring managerial approval; all overrides are logged. Monitoring ownership sits with the pricing steward and MLOps, with alerts routed via the production monitoring dashboard.

\subsection{Risk Envelope Testing}

We evaluate the risk management capabilities through stress testing under extreme scenarios: economic downturn (20\% demand reduction), competitor price war (15\% competitor price cut), and cost inflation (25\% infrastructure cost increase). Figure 2 illustrates the performance comparison across pricing strategies under these adverse conditions.

\begin{figure}[!t]
\centering
\begin{tikzpicture}
\begin{scope}[xscale=1.2, yscale=0.8]
\draw[->] (0,0) -- (5,0) node[right] {Scenario};
\draw[->] (0,0) -- (0,5) node[above] {Performance Index};

\draw[thick, blue] (0.5,4.5) .. controls (1.5,4.2) and (2.5,3.8) .. (4.5,2.8);
\node[blue] at (4.7,2.5) {Static};

\draw[thick, red] (0.5,4.8) .. controls (1.5,4.0) and (2.5,2.8) .. (4.5,1.5);
\node[red] at (4.7,1.2) {Uniform};

\draw[thick, green] (0.5,4.3) .. controls (1.5,4.1) and (2.5,3.9) .. (4.5,3.7);
\node[green] at (4.7,3.4) {Guardrailed};

\draw[dashed] (1.5,0) -- (1.5,5);
\draw[dashed] (3.0,0) -- (3.0,5);
\draw[dashed] (4.5,0) -- (4.5,5);

\node at (1.5,-0.3) {Mild};
\node at (3.0,-0.3) {Moderate};
\node at (4.5,-0.3) {Severe};
\end{scope}
\end{tikzpicture}
\caption{Risk Envelope Performance Under Adverse Scenarios}
\label{fig:risk_envelope}
\end{figure}
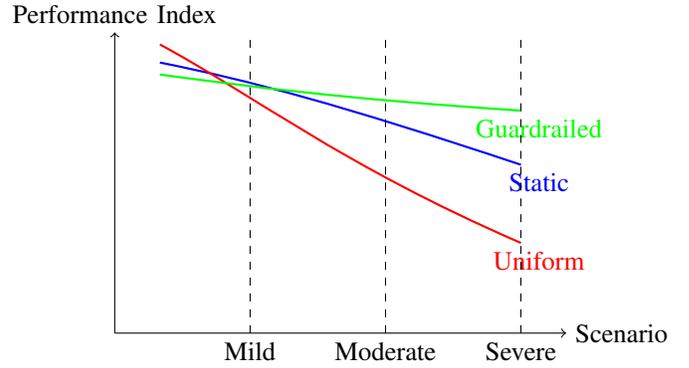

The guardrailed approach demonstrates significantly better resilience under stress conditions, maintaining positive performance even in severe scenarios where alternative strategies deteriorate rapidly. This robustness stems from the constraint structure that prevents excessive risk-taking and the probabilistic forecasting that anticipates adverse conditions.

The risk envelope testing also validates the early warning capabilities of the system, with alert triggers occurring 2-3 weeks before material business impact in simulated scenarios. This lead time provides management with opportunity to implement mitigating actions, transforming pricing from a reactive to proactive business function.

\section{Validation Protocol and Key Metrics}

\textbf{Data Scope and Splits.} 
This data set contains 2.3 million subscription records from three SaaS domains over 18 months. To prevent temporal leakage, we separate by time, using the first 12 months for training, the next 3 for validation, and the last 3 for testing. The characteristics of a prediction point are estimated using only past data available before that point in time.

\textbf{Evaluation Protocol.}
We utilize rolling-window backtesting to measure 30-day forecasts. For each month in the test period, we retrain models on all data until that time, generate price recommendations, and simulate what would have happened against held-out actuals. This mimics the actual deployment of models, which is periodically refreshed.

\textbf{Key Metrics Reported.}
We evaluate across four dimensions:
\begin{itemize}
    \item \textbf{Forecast Accuracy:} MAPE (2.18\%), RMSE (2.57), Interval Coverage Probability (95\%)
    \item \textbf{Business Impact:} Revenue lift (+16.2\%), margin improvement (+14.3\%), churn rate (2.3\%), CLV change (+11.4\%).
    \item \textbf{Constraint Compliance:} 99.7\% of recommendations satisfied all guardrails; violations triggered automated fallback to safe defaults
    \item \textbf{Scalability:} Median optimization runtime 4.3 s (p95 8.1 s) for 500 segments on 16 vCPU cloud instances.
\end{itemize}

\textbf{Conclusion.}
The guardrailed elasticity pricing framework provides a comprehensive approach to subscription strategy that balances revenue optimization with customer relationship preservation. By combining advanced analytics with business constraints and ethical safeguards, it enables sustainable growth in increasingly competitive digital markets. Organizations adopting this approach position themselves for leadership in the subscription economy through superior monetization, enhanced resilience, and strengthened customer trust.

\end{document}